# Rollout Sampling Policy Iteration for Decentralized POMDPs

**Feng Wu**
School of Computer Science
University of Sci. & Tech. of China
Hefei, Anhui 230027 China
wufeng@mail.ustc.edu.cn

**Shlomo Zilberstein**
Department of Computer Science
University of Massachusetts
Amherst, MA 01003 USA
shlomo@cs.umass.edu

**Xiaoping Chen**
School of Computer Science
University of Sci. & Tech. of China
Hefei, Anhui 230027 China
xpchen@ustc.edu.cn

## Abstract

We present *decentralized rollout sampling policy iteration* (DecRSPI) — a new algorithm for multi-agent decision problems formalized as DEC-POMDPs. DecRSPI is designed to improve scalability and tackle problems that lack an explicit model. The algorithm uses Monte-Carlo methods to generate a sample of reachable belief states. Then it computes a joint policy for each belief state based on the rollout estimations. A new policy representation allows us to represent solutions compactly. The key benefits of the algorithm are its linear time complexity over the number of agents, its bounded memory usage and good solution quality. It can solve larger problems that are intractable for existing planning algorithms. Experimental results confirm the effectiveness and scalability of the approach.

## 1 Introduction

Planing under uncertainty in multi-agent settings is a challenging computational problem, particularly when agents with imperfect sensors and actuators, such as autonomous rovers or rescue robots, must reason about a large space of possible outcomes and choose a plan based on their incomplete knowledge. The *partially observable Markov decision process* (POMDP) has proved useful in modeling and analyzing this type of uncertainty in single-agent domains. When multiple cooperative agents are present, each agent must also reason about the decisions of the other agents and how they may affect the environment. Since each agent can only obtain partial information about the environment and sharing all the local information among the agents is often impossible, each agent must act based solely on its local information. These problems can be modeled as *decentralized POMDPs* (DEC-POMDPs) [2].

When a complete model of the domain is available, DEC-POMDPs can be solved using a wide range of optimal or approximate algorithms, particularly MBDP [18] and its descendants [1, 8, 17]. Unfortunately, these algorithms are quite limited in terms of the size of the problems they can tackle. This is not surprising given that finite-horizon DEC-POMDPs are NEXP-complete [2]. Intuitively, the main reason is that it is hard to define a compact belief state and compute a value function for DEC-POMDPs, as is often done for POMDPs. The state and action spaces blow-up exponentially with the number of agents. Besides, it is very difficult to search over the large policy space and find the best action for every possible situation.

Another key challenge is modeling the dynamics of the entire domain, which may include complex physical systems. Existing DEC-POMDP algorithms assume that a complete model of the domain is known. This assumption does not hold in some real-world applications such as robot soccer. Incomplete domain knowledge is often addressed by reinforcement learning algorithms [19]. However, most cooperative multi-agent reinforcement learning algorithms assume that the system state is completely observable by all the agents [6]. Learning cooperative policies for multi-agent partially-observable domains is extremely challenging due to the large space of possible policies given only the local view of each agent.

In reinforcement learning, a class of useful techniques such as Monte-Carlo methods allows agents to choose actions based on experience [19]. These methods require no prior knowledge of the dynamics, as long as sample trajectories can be generated online or using a simulator of the environment. Although a model is required, it must only provide enough information to generate samples, not the complete probability distributions of all possible transitions that are required by planning algorithms. In many cases it is easy to generate samples by simulating the target environment, but obtaining distributions in explicit form may be much harder. In the robot soccer domain, for example, there exist many high-fidelity simulation engines. It is also possible to put a central camera on top of the field and obtain samples by running the actual robots.

This paper introduces the *decentralized rollout sampling*

*policy iteration* (DecRSPI) algorithm for finite-horizon DEC-POMDPs. Our objective is to compute a set of cooperative policies using Monte-Carlo methods, without having an explicit representation of the dynamics of the underlying system. DecRSPI first samples a set of reachable belief states based on some heuristic policies. Then it computes a joint policy for each belief state based on the rollout estimations. Similar to dynamic programming approaches, policies are constructed from the last step backwards. A new policy representation is used to bound the amount of memory. To the best of our knowledge, this is the first rollout-based learning algorithm for finite-horizon DEC-POMDPs. DecRSPI has linear time complexity over the number of agents and it can solve much larger problems compared to existing planning algorithms.

We begin with some background on the DEC-POMDP model and the policy structure we use. We then describe each component of the rollout sampling algorithm and analyze its properties. Finally, we examine the performance of DecRSPI on a set of test problems, and conclude with a summary of related work and the contributions.

## 2 Decentralized POMDPs

Formally, a finite-horizon DEC-POMDP can be defined as a tuple $\langle I, S, \{A_i\}, \{\Omega_i\}, P, O, R, b^0, T \rangle$, where

- $I$ is a collection of agents, identified by $i \in \{1 \ldots m\}$, and $T$ is the time horizon of the problem.
- $S$ is a finite state space and $b^0$ is the initial belief state (i.e., a probability distribution over states).
- $A_i$ is a discrete action space for agent $i$. We denote by $\vec{a} = \langle a_1, a_2, \cdots, a_m \rangle$ a joint action where $a_i \in A_i$ and $\vec{A} = \times_{i \in I} A_i$ is the joint action space.
- $\Omega_i$ is a discrete observation space for agent $i$. Similarly $\vec{o} = \langle o_1, o_2, \cdots, o_m \rangle$ is a joint observation where $o_i \in \Omega_i$ and $\vec{\Omega} = \times_{i \in I} \Omega_i$ is the joint observation space.
- $P : S \times \vec{A} \to \Delta(S)$ is the state transition function and $P(s'|s, \vec{a})$ denotes the probability of the next state $s'$ when the agents take joint action $\vec{a}$ in state $s$.
- $O : S \times \vec{A} \to \Delta(\vec{\Omega})$ is an observation function and $O(\vec{o}|s', \vec{a})$ denotes the probability of observing $\vec{o}$ after taking joint action $\vec{a}$ with outcome state $s'$.
- $R : S \times \vec{A} \to \mathcal{R}$ is a reward function and $R(s, \vec{a})$ is the immediate reward after agents take $\vec{a}$ in state $s$.

In a DEC-POMDP, each agent $i \in I$ executes an action $a_i$ based on its policy at each time step $t$. Thus a joint action $\vec{a}$ of all the agents is performed, followed by a state transition of the environment and an identical joint reward obtained by the team. Then agent $i$ receives its private observation $o_i$ from the environment and updates its policy for the next execution cycle. The goal of each agent is to choose a policy that maximizes the accumulated reward of the team over the horizon $T$, i.e. $\sum_{t=1}^{T} E[R(t)|b^0]$.

Generally, a policy $q_i$ is a mapping from agent $i$'s observation history to an action $a_i$ and a joint policy $\vec{q} = \langle q_1, q_2, \cdots, q_m \rangle$ is a vector of policies, one for each agent. The value of a fixed joint policy $\vec{q}$ at state $s$ can be computed recursively by the Bellman equation:

$$V(s, \vec{q}) = R(s, \vec{a}) + \sum_{s', \vec{o}} P(s'|s, \vec{a}) O(\vec{o}|s', \vec{a}) V(s', \vec{q}_{\vec{o}})$$

where $\vec{a}$ is the joint action specified by $\vec{q}$, and $\vec{q}_{\vec{o}}$ is the joint sub-policy of $\vec{q}$ after observing $\vec{o}$. Given a state distribution $b \in \Delta(s)$, the value of a joint policy $\vec{q}$ can be computed by

$$V(b, \vec{q}) = \sum_{s \in S} b(s) V(s, \vec{q}) \quad (1)$$

Note that in a DEC-POMDP, each agent can only receive its own local observations when executing the policy. Therefore the policy must be completely decentralized, which means the policy of an agent must be guided by its own local observation history only. It is not clear how to maintain a sufficient statistic, such as a *belief state* in POMDPs, based only on the local partial information of each agent. Thus, most of the works on multi-agent partially observable domains are policy-based and learning in DEC-POMDP settings is extremely challenging. While the policy execution is decentralized, planning or learning algorithms can operate offline and thus may be centralized [11, 18].

The policies for finite-horizon DEC-POMDPs are often represented as a set of local *policy trees* [1, 8, 11, 17, 18]. Each tree is defined recursively with an action at the root and a subtree for each observation. This continues until the horizon is reached at a leaf node. A *dynamic programming* (DP) algorithm was developed to build the policy trees optimally from the bottom up [11]. In this algorithm, the policies of the next iteration are enumerated by an *exhaustive backup* of the current trees. That is, for each action and each resulting observation, a branch to any of the current trees is considered. Unfortunately, the number of possible trees grows double-exponentially over the horizon. Recently, memory-bounded techniques have been introduced. These methods keep only a fixed number of trees at each iteration [1, 8, 17, 18]. They use a fixed amount of memory and have linear complexity over the horizon.

There are many possibilities for constructing policies with bounded memory. In this work we use a *stochastic policy* for each agent. It is quite similar to stochastic *finite state controllers* (FSC), used to solve infinite-horizon POMDPs [16] and DEC-POMDPs [3]. But our stochastic policies have a layered structure, one layer for each time step. Each layer has a fixed number of decision nodes. Each node is labeled with an action and includes a node selection function. The selection function is a mapping from an observation to a probability distribution over the nodes of the next layer. In this paper, we denote by $Q_i^t$ the set of decision nodes of agent $i \in I$ at time step $t \in 1..T$. Also, $N$ denotes the predetermined size of $Q_i^t$ and $\pi(q_i'|o_i)$ is the probability of selecting the node of the next layer $q_i'$ after observing $o_i$.

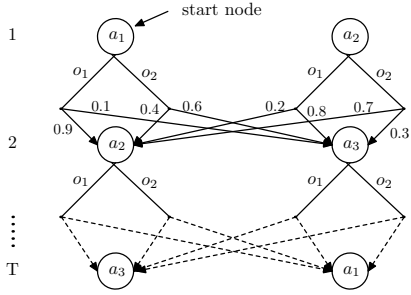

Figure 1: An agent's stochastic policy with two observations and two decision nodes in each layer.

An example of such stochastic policies is shown in Figure 1. In the planning phase, a set of stochastic polices are constructed offline, one for each agent. When executing the policy, each agent executes the action in the current node and then transitions to the next node based on the received observation as well as the node selection function. We show how the stochastic node selection function can be optimized easily by our policy improvement technique. The following sections describe the algorithm in details.

## 3 The Rollout Sampling Method

In this section, we propose a new *rollout sampling policy iteration* (DecRSPI) for DEC-POMDPs that heuristically generates stochastic policies using an approximate policy improvement operator trained with Monte-Carlo simulation. The approximate operator performs policy evaluation by simulation, evaluating a joint policy $\vec{q}$ at state $s$ by drawing $K$ sample trajectories of $\vec{q}$ starting at $s$. Then, the operator performs policy improvement by constructing a series of linear programs with parameters computed from samples and then solving the linear programs to induce a new improved approximate policy. Similar to MBDP, DecRSPI generates policies using point-based dynamic programming, which builds policies according to heuristic state distributions from the bottom up. The key difference is that DecRSPI improves the policies by simulation without knowing the exact transition function $P$, observation function $O$ and reward function $R$ of the DEC-POMDP model.

Note that DecRSPI is performed offline in a centralized way, but the computed policies are totally decentralized. The use of simulation assumes that the state of the environment can be reset and the system information (state, reward and observations) are available after executing a joint action by the agents. In the planning phase, this information is often available. In large real-world systems, modeling

**Algorithm 1**: Rollout Sampling Policy Iteration

generate a random joint policy $\vec{Q}$ given $T, N$
sample a set of beliefs $B$ for $t \in 1..T, n \in 1..N$
**for** $t=T$ to $1$ **do**
    **for** $n=1$ to $N$ **do**
        $b \leftarrow B_n^t, \vec{q} \leftarrow \vec{Q}_n^t$
        **repeat**
            **foreach** *agent* $i \in I$ **do**
                keep the other agents' policies $q_{-i}$ fixed
                **foreach** *action* $a_i \in A_i$ **do**
                      $\Phi_i \leftarrow$ estimate the parameter matrix
                      build a linear program with $\Phi_i$
                      $\pi_i \leftarrow$ solve the linear program
                      $\Delta_i \leftarrow \Delta_i \cup \{\langle a_i, \pi_i \rangle\}$
                $\langle a_i, \pi_i \rangle^* \leftarrow \arg\max_{\Delta_i}$ **Rollout**$(b, \langle a_i, \pi_i \rangle)$
                update agent $i$'s policy $q_i$ by $\langle a_i, \pi_i \rangle^*$
        **until** *no improvement in all agents' policies*

**return** the joint policy $\vec{Q}$

the exact DEC-POMDP is extremely challenging and even the representation itself is nontrivial for several reasons. First, the system may be based on some complex physical models and it may be difficult to compute the exact $P$, $O$ and $R$. Second, the state, action and observation spaces may be very large, making it hard to store the entire transition table. Fortunately, simulators of these domains are often available and can be modified to compute the policies as needed.

We activate DecRSPI by providing it with a random joint policy and a set of reachable state distributions, computed by some heuristics. The joint policy is initialized by assigning a random action and random node selection functions for each decision node from layer-1 to layer-$T$. Policy iteration is performed from the last step $t=T$ backward to the first step $t=1$. At each iteration, we first choose a state distribution and an unimproved joint policy. Then we try to improve the joint policy based on the state distribution. This is done by keeping the policies of the other agents fixed and searching for the best policy of one agent at a time. We continue to alternate between the agents until no improvement is achievable for the current policies of all the agents. This process is summarized in Algorithm 1.

### 3.1 Belief Sampling

In this paper, the *belief state* $b \in \Delta(S)$ is a probability distribution over states. We use it interchangeably with the *state distribution* with the same meaning. Generally, given belief state $b^t$ at time $t$, we determine $\vec{a}^t$, execute $\vec{a}^t$ and make a subsequent observation $\vec{o}^{t+1}$, then update our belief state to obtain $b^{t+1}$. In single-agent POMDPs, this belief state is obtained via straightforward Bayesian updating, by computing $b^{t+1} = Pr(S|b^t, \vec{a}^t, \vec{o}^{t+1})$. Unfortunately, even if the transition and observation functions are available, the belief update itself is generally time-consuming

**Algorithm 2**: Belief Sampling

**for** *n=1* to *N* **do**
    $B_n^t \leftarrow \emptyset$ for $t \in 1..T$
    $h \leftarrow$ choose a heuristic policy
    **for** *k=1* to *K* **do**
        $s \leftarrow$ draw a state from $b^0$
        **for** *t=1* to *T* **do**
            $\theta^t \leftarrow \theta^t \cup \{b_k^t(s)\}$
            $\vec{a} \leftarrow$ select a joint action based on $h$
            $s' \leftarrow$ simulate the model with $s, \vec{a}$
            $s \leftarrow s'$
    **for** *t=1* to *T* **do**
        $b^t \leftarrow$ compute the belief by particle set $\theta^t$
        $B_n^t \leftarrow B_n^t \cup \{b^t\}$
**return** the belief set $B$

---

because each belief state is a vector of size $|S|$. To approximate belief states by simulation, consider the following particle filtering procedure. At any time step $t$, we have a collection $\theta^t$ of $K$ particles. The particle set $\theta^t, t \in 1..T$ represents the following state distribution:

$$b^t(s) = \frac{\sum_{k=1}^{K}\{1 : b_k^t(s) \in \theta^t\}}{K}, \forall s \in S \qquad (2)$$

where $b_k^t(s)$ is the $k^{th}$ particle of $\theta^t$. As mentioned above, the significance of this method lies in the fact that, for many applications, it is easy to sample successor states according to the system dynamics. But direct computation of beliefs is generally intractable especially when the dynamics specification is unavailable.

Another key question is how to choose the heuristic policies. In fact, the usefulness of the heuristics and, more importantly, the computed belief states, are highly dependent on the specific problem. Instead of just using one heuristic, a whole portfolio of heuristics can be used to compute a set of belief states. Thus, each heuristic is used to select a subset of the policies. There are a number of possible alternatives. Our first choice is the random policy, where agents select actions randomly from a uniform distribution at each time step. Another choice is the policy of the underlying MDP. That is, agents can learn an approximate MDP value function by some MDP learning algorithms and then select actions greedily based on that value function. In specific domains such as robot soccer, where learning the MDP policy is also hard, hand-coded policies or policies learned by DecRSPI itself with merely random guidance are also useful as heuristics. The overall belief sampling method is detailed in Algorithm 2.

### 3.2 Policy Improvement

In multi-agent settings, agents with only local information must reason about all the possible choices of the others and select the optimal joint policy that maximizes the team's expected reward. One straightforward method for finding

Maximize$_x \sum_{o_i \in \Omega_i} \sum_{q'_i \in Q_i^{t+1}} \Phi_i(o_i, q'_i) x(o_i, q'_i)$
subject to $\forall_{o_i, q'_i} x(o_i, q'_i) \geq 0, \forall_{o_i} \sum_{q'_i} x(o_i, q'_i) = 1.$

Table 1: Linear program to improve agent $i$'s policy where the variable $x(o_i, q'_i) = \pi(q'_i|o_i)$ is the node selection table.

the optimal joint policy is to simply search over the entire space of possible policies, evaluate each one, and select the policy with the highest value. Unfortunately, the number of possible joint policies is $\mathcal{O}((|A_i|^{(|\Omega_i|^T-1)/(|\Omega_i|-1)})^{|I|})$. Instead of searching over the entire policy space, dynamic programming (DP) constructs policies from the last step up to the first one and eliminates dominated policies at the early stages [11]. However, the *exhaustive backup* in the DP algorithm at $t$ still generates agent $i$'s policies of the order $\mathcal{O}(|A_i||Q_i^{t-1}|^{|\Omega_i|})$. Memory-bounded techniques have been developed to combine the top-down heuristics and the bottom-up dynamic programming together, keeping only a bounded number of policies at each iteration [18]. This results in linear complexity over the horizon, but the one-step backup operation is still time-consuming [17].

Our algorithm is based on the MBDP algorithm [18], but it approximates the backup operation with an alternating maximization process. As shown in Algorithm 1, the basic idea is to choose each agent in turn and compute the best-response policy, while keeping the policies of the other agents fixed. This process is repeated until no improvement is possible for all agents. That is, the process ends when the joint policy converges to a Nash equilibrium. This method was first introduced by Nair et al. [14] and later refined by Bernstein et al. [3]. The differences are: Nair et al. use the method to reformulate the problem as an augmented POMDP; Bernstein et al. use it to optimize the controllers of infinite-horizon DEC-POMDPs. In contrast, when an agent is chosen, our algorithm approximates the best-response policy that maximizes the following value:

$$V(b, \vec{q}) = R(b, \vec{a}) + \sum_{s', \vec{o}, \vec{q}'} Pr(s', \vec{o}|b, \vec{a}) \prod_i \pi(q'_i|o_i) V(s', \vec{q}')$$
(3)

where $Pr(s', \vec{o}|b, \vec{a}) = \sum_{s \in S} b(s) P(s'|s, \vec{a}) O(\vec{o}|s', \vec{a})$ and $R(b, \vec{a}) = \sum_{s \in S} b(s) R(s, \vec{a})$. This value function is similar to Equation 1, but for a stochastic joint policy.

Notice that our algorithm is designed to work when an explicit form of system dynamics is not available. Our solution, as shown in Algorithm 1, is two-fold: first we find the best node selection function $\pi_i$ for every action $a_i \in A_i$ and generate a set of stochastic policies $\Delta_i$; then we evaluate the policy $q_i \in \Delta_i$ for the given belief point $b$, choose the best one and update the current policy of agent $i$ with it. In order to find the best $\pi_i$ that maximizes the value function of Equation 3 given $a_i$ and other agents' policies $q_{-i}$, we use the linear program shown in Table 1. Note that

**Algorithm 3**: Parameter Estimation

**Input**: $b, a_i, q_{-i}$
$a_{-i} \leftarrow$ get actions from $q_{-i}$
**for** $k=1$ **to** $K$ **do**
    $s \leftarrow$ draw a state from $b$
    $s', \vec{o} \leftarrow$ simulate the model with $s, \vec{a}$
    $\omega_{o_i}(s', o_{-i}) \leftarrow \omega_{o_i}(s', o_{-i}) + 1$
normalize $\omega_{o_i}$ for $\forall o_i \in \Omega_i$
**foreach** $o_i \in \Omega_i, q'_i \in Q_i^{t+1}$ **do**
    **for** $k=1$ **to** $K$ **do**
        $s', o_{-i} \leftarrow$ draw a sample from $\omega_{o_i}$
        $q'_{-i} \leftarrow$ get other agents' policy $\pi(\cdot|q_{-i}, o_{-i})$
        $\Phi_i(o_i, q'_i)_k \leftarrow$ **Rollout**$(s', \vec{q}')$
    $\Phi_i(o_i, q'_i) \leftarrow \frac{1}{K} \sum_{k=1}^{K} \Phi_i(o_i, q'_i)_k$
**return** the parameter matrix $\Phi_i$

$R(b, \vec{a})$ is a constant given $b, a_i, q_{-i}$ and is thus omitted. The matrix $\Phi_i$ of the linear program is defined as follows:

$$\Phi_i(o_i, q'_i) = \sum_{s', o_{-i}, q'_{-i}} Pr(s', \vec{o}|b, \vec{a}) \pi(q'_{-i}|o_{-i}) V(s', \vec{q}')$$

where $\pi(q'_{-i}|o_{-i}) = \prod_{k \neq i} \pi(q'_k|o_k)$. Since the dynamics is unknown, Algorithm 3 is used to estimate $\Phi_i$. It first estimates $Pr(s', \vec{o}|b, \vec{a})$ by drawing $K$ samples from one-step simulation. Then it estimates each element of $\Phi_i$ by another $K$ samples with $\pi(q'_{-i}|o_{-i})$. The value of $V(s', \vec{q}')$ is approximated by the rollout operation as follows.

### 3.3 Rollout Evaluation

The rollout evaluation is a Monte-Carlo method to estimate the value of a policy $\vec{q}$ at a state $s$ (or belief state $b$), without requiring an explicit representation of the value function as the DP algorithm does. A rollout for $\langle s, \vec{q} \rangle$ simulates a trajectory starting from state $s$ and choosing actions according to policy $\vec{q}$ up to the horizon $T$. The observed total accumulated reward is averaged over $K$ rollouts to estimate the value $V(s, \vec{q})$. If a belief state $b$ is given, it is straightforward to draw a state $s$ from $b$ and perform this simulation. The outline of the rollout process is given in Algorithm 4.

The accuracy of the expected value estimate improves with the number of rollouts. Intuitively, the value starting from $\langle s, \vec{q} \rangle$ can be viewed as a random variable whose expectation is $V(s, \vec{q})$. Each rollout term $v_k$ is a sample of this random variable and the average of these $\tilde{V}$ is an unbiased estimate of $V(s, \vec{q})$. Thus, we can apply the following Hoeffding bounds to determine the accuracy of this estimate.

**Property 1** (Hoeffding inequality). *Let $V$ be a random variable in $[V_{\min}, V_{\max}]$ with $\bar{V} = E[V]$, observed values $v_1, v_2, \cdots, v_K$ of $V$, and $\tilde{V} = \frac{1}{K} \sum_{k=1}^{K} v_k$. Then*

$$Pr(\tilde{V} \leq \bar{V} + \varepsilon) \geq 1 - \exp\left(-2K\varepsilon^2/V_\Delta^2\right)$$
$$Pr(\tilde{V} \geq \bar{V} - \varepsilon) \geq 1 - \exp\left(-2K\varepsilon^2/V_\Delta^2\right)$$

where $V_\Delta = V_{\max} - V_{\min}$ is the range of values.

**Algorithm 4**: Rollout Evaluation

**Input**: $t, s, \vec{q}^t$
**for** $k=1$ **to** $K$ **do**
    $v_k \leftarrow 0$
    **while** $t \leq T$ **do**
        $\vec{a} \leftarrow$ get the joint action from $\vec{q}^t$
        $s', r, \vec{o} \leftarrow$ simulate the model with $s, \vec{a}$
        $v_k \leftarrow v_k + r, \vec{q}^{t+1} \leftarrow \pi(\cdot | \vec{q}^t, \vec{o})$
        $s \leftarrow s', t \leftarrow t + 1$
$\tilde{V} \leftarrow \frac{1}{K} \sum_{k=1}^{K} v_k$
**return** the average value $\tilde{V}$

Given a particular confidence threshold $\delta$ and a size of samples $K$, we can produce a PAC-style error bound $\varepsilon$ on the accuracy of our estimate $\tilde{V}$:

$$\varepsilon = \sqrt{\frac{V_\Delta^2 \ln(\frac{1}{\delta})}{2K}} \quad (4)$$

**Property 2.** *If the number of rollouts $K$ is infinitely large, the average value returned by the rollout algorithm $\tilde{V}$ will converge to the expected value of the policy $\bar{V}$.*

The required sample size given error tolerance $\varepsilon$ and confidence threshold $\delta$ for the estimation of $\tilde{V}$ is:

$$K(\varepsilon, \delta) = \frac{V_\Delta^2 \ln(\frac{1}{\delta})}{2\varepsilon^2} \quad (5)$$

It is difficult to compute a meaningful error bound for the overall algorithm. There are several reasons: (1) DecRSPI is an MBDP-based algorithm and MBDP itself has no guarantee on the solution quality since the belief sampling method is based on domain-dependent heuristics; (2) the local search technique — which updates one agent's policy at a time — could get stuck in a suboptimal Nash equilibrium; and (3) the error may accumulate over the horizon, because the policies of the current iteration depend on the policies of previous iterations. Thus, we demonstrate the performance and benefits of DecRSPI largely based on experimental results.

### 3.4 Complexity Analysis

Note that the size of each agent's policy is predetermined with $T$ layers and $N$ decision nodes in each layer. At each iteration, DecRSPI chooses an unimproved joint policy and tries to improve the policy parameters (actions and node selection functions) of each agent. Thus, the amount of space is of the order $\mathcal{O}(mTN)$ for $m$ agents. Several rollouts are performed in the main process of each iteration. The time per rollout grows linearly with $T$. Therefore, the total time with respect to the horizon is on the order of $1 + 2 + \cdots + T = (T^2 + T)/2$, i.e. $\mathcal{O}(T^2)$.

**Theorem 3.** *The DecRSPI algorithm has linear space and quadratic time complexity with respect to the horizon $T$.*

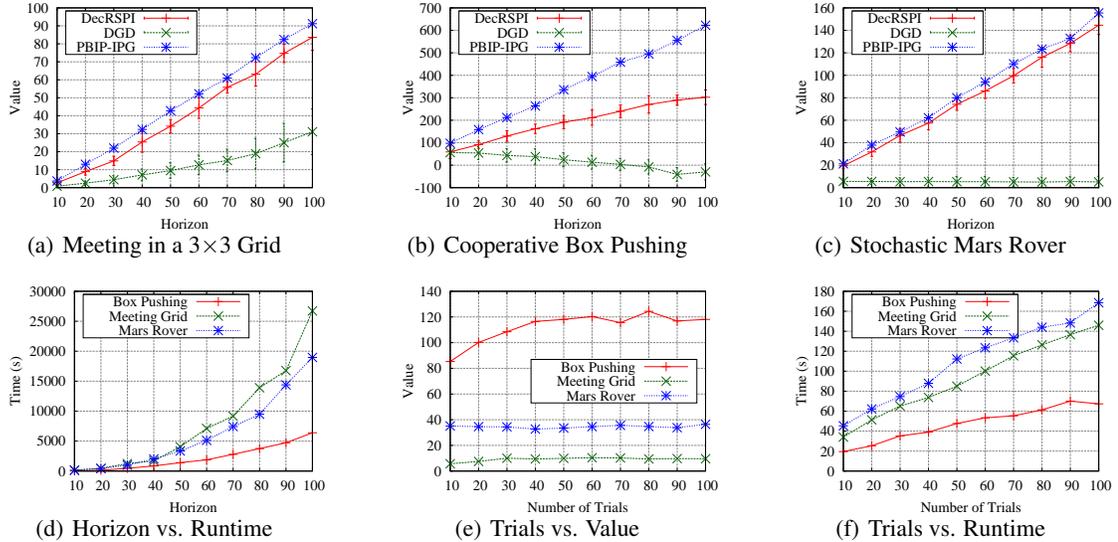

Figure 2: Experimental results for standard benchmark problems.

Clearly, the amount of space grows linearly with the number of agents. At each iteration, the main loop chooses $N$ joint policies. For each joint policy, the improvement process selects agents alternatively until no improvement is possible. In practice, we set thresholds both for the minimum improvement (e.g. $10^{-4}$) and the maximum repeat count (e.g. 100). The improvement process terminates when one of these bounds is reached. Theoretically, the runtime of a rollout inside the improvement process is independent of the number of agents. However in practice, systems with more agents will take significantly more time to simulate, thereby increasing the time per rollout. But this is due to the complexity of domains or simulators, not the complexity of the DecRSPI algorithm.

**Theorem 4.** *Ignoring system simulation time, the DecRSPI algorithm has linear time and space complexity with respect to the number of agents $|I|$.*

## 4 Experiments

We performed experiments on several common benchmark problems in the DEC-POMDP literature to evaluate the solution quality and runtime of DecRSPI. A larger distributed sensor network domain was used to test the scalability of DecRSPI with respect to the number of agents.

### 4.1 Benchmark Problems

We first tested DecRSPI on several common benchmark problems for which the system dynamics — an explicit representation of the transition, observation and reward functions — is available. To run the learning algorithm, we implemented a DEC-POMDP simulator based on the dynamics and learned the joint policy from the *simulator*. We used two types of heuristic policies to sample belief states: the random policy that randomly chooses an action with a uniform distribution, and the MDP-based policy that chooses an action according to the global state (which is known during the learning phase). For the benchmark problems, we solved the underlying MDP models and used the policies for sampling. DecRSPI selects a heuristic each time with a chance of 0.55 acting randomly and 0.45 for MDP policies.

There are few work on learning policies in the general DEC-POMDP setting. In the experiments, we compared our results with the *distributed gradient descent* (DGD) [15] with different horizons. The DGD approach performs the gradient-descent algorithm for each agent independently to adapt the parameters of each agent's local policy. We also present the results of PBIP-IPG [1] — the best existing planning algorithm — for these domains. Notice that PBIP-IPG computes the policy based on an explicit model of system dynamics. Thus, the values of PBIP-IPG can be viewed as upper bounds for learning algorithms. Due to the randomness of Monte-Carlo methods, we ran the algorithm 20 times per problem and reported average runtimes and values. The default number of policy nodes $N$ is 3 and the number of samples $K$ is 20.

We experimented with three common DEC-POMDP benchmark problems, which are also used by PBIP-IPG [1]. The Meeting in a 3×3 Grid problem [3] involves two robots that navigate in a 3×3 grid and try to stay as much time as possible in the same cell. We adopted the version used by Amato et al. [1], which has 81 states, 5 actions and 9 observations per robot. The results for this domain with different horizons are given in Figure 2(a). The Cooperative Box Pushing problem [17] involves two robots that cooperate with each other to push boxes to their destinations in a 3×4 grid. This domain has 100 states, 4 actions and 5 observations per robot. The results are given in Figure 2(b). The Stochastic Mars Rover problem [1] is a larger domain with 2 robots, 256 states, 6 actions and 8 observations per robot. The results are given in Figure 2(c).

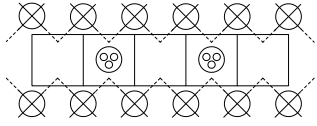

Figure 3: The distributed sensor network domain.

In all three domains, DecRSPI outperforms DGD with large margins. In the Meeting in Grid and Mars Rover domains, the learning results of DecRSPI are quite close to the planning values of PBIP-IPG. Note that PBIP-IPG is also an MBDP-based algorithm whose values represent good upper bounds on the learning quality of DecRSPI. Being close to the value of PBIP-IPG means that DecRSPI does learn good policies given the same heuristics and policy sizes. In the Cooperative Box Pushing domain, the gap between DecRSPI and PBIP-IPG is a little bit larger because this problem has more complex interaction structure than the other two domains. Interestingly, in this domain, the value of DGD decreases with the horizon.

We also present timing results for each domain with different horizons ($T$) in Figure 2(d), which shows the same property (quadratic time complexity) as stated in Theorem 3. In Figure 2(e), we test DecRSPI with different number of trials ($K$) and a fixed horizon of 20. The value of the Meeting in Grid and Mars Rover domains becomes stable when the number of trials is larger than 10. But the Box Pushing problem needs more trials (about 40) to get to a stable value, which is very close to the value of PBIP-IPG. In Figure 2(f), we show that runtime grows linearly with the number of trials in all three domains. It is worthwhile to point out that in these experimental settings, DecRSPI runs much faster than PBIP-IPG. For example, in the Stochastic Mars Rover domain with horizon 20, PBIP-IPG may take 14947s while DecRSPI only needs 49.8s.

### 4.2 Distributed Sensor Network

The distributed sensor network (DSN) problem, adapted from [20], consists of two chains with identical number of sensors as shown in Figure 3. The region between the chains is split into cells and each cell is surrounded by four sensors. The two targets can move around in the place, either moving to a neighboring cell or staying in place with equal probability. Each target starts with an energy level of 2. A target is captured and removed when it reaches 0. Each sensor can take 3 actions ( track-left, track-right and none) and has 4 observations (left and right cells are occupied or not), resulting in joint spaces of $3^{|I|}$ actions and $4^{|I|}$ observations (e.g. $3^{20} \approx 3.5 \times 10^9$ joint actions and $4^{20} \approx 1.1 \times 10^{12}$ joint observations for the 20 agents case). Each track action has a *cost* of 1. The energy of a target will be decreased by 1 if it is tracked by at least three of the four surrounding sensors at the same time. When a target is captured, the team gets a reward of 10. When all targets are captured, the DSN restarts with random target positions.

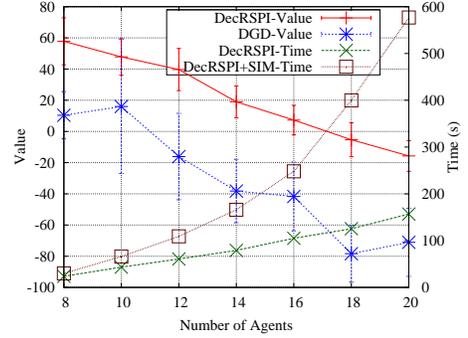

Figure 4: Value and runtime of DSN ($T$=10, $N$=3, $K$=20).

This domain is designed to demonstrate the scalability of DecRSPI over the number of agents. Most of the planning algorithms for general DEC-POMDPs have only been tested in domains with 2 agents [1, 8, 11, 17, 18]. It is very challenging to solve a general DEC-POMDP problems with many agents because the joint action and observation spaces grow *exponentially* over the number of agents. Our results in the DSN domain with horizon 10 and random heuristics are shown in Figure 4. Again, DecRSPI achieves much better value than DGD. The value decreases with the growing number of agents because there are more cells for the targets to move around and greater chance of sensor miscoordination. Note that DecRSPI solves this problem without using any specific domain structure and the learned policies are totally decentralized, without any assumption of communication or global observability. The figure also shows two measures of timing results: DecRSPI-Time — the runtime of DecRSPI, and DecRSPI+SIM-Time — the overall runtime including domain simulation time. The two measures of runtime grow with the number of agents. As stated in Theorem 4, DecRSPI-Time grows linearly, which shows that it scales up very well with the number of agents.

## 5 Related Work

Several policy search algorithms have been introduced to learn agents' policies without the system dynamics. The *distributed gradient descent* (DGD) algorithm performs gradient-based policy search *independently* on each agent's local controller using the experience data [15]. Zhang et al. [20] proposed an online natural actor-critic algorithm using *conditional random fields* (CRF). It can learn cooperative policies with CRFs, but it assumes that agents can communicate and share their local observations at every step. Melo [13] proposed another actor-critic algorithm with natural gradient, but it only works for *transition-independent DEC-POMDPs*. In contract, our algorithm learns cooperative policies for the general DEC-POMDP setting without any assumption about communication.

The *rollout sampling* method has been introduced to learn MDP policies without explicitly representing the value function [9, 10, 12]. The main idea is to produce training data through extensive simulation (rollout) of the previous policy and use a supervised learning algorithm (e.g. SVM) to learn a new policy from the labeled data. The *rollout* technique is also widely used to perform lookahead and estimate the value of action in online methods [4, 5, 7]. Our algorithm uses *rollout sampling* to estimate the parameters of policy improvements and select the best joint policy.

## 6 Conclusion

We have presented the *decentralized rollout sampling policy iteration* (DecRSPI) algorithm for learning cooperative policies in partially observable multi-agent domains. The main contribution is the ability to compute decentralized policies without knowing explicitly the system dynamics. In many applications, the system dynamics is either too complex to be modeled accurately or too large to be represented explicitly. DecRSPI learns policies from experience obtained by merely interacting with the environment. The learned policies are totally decentralized without any assumption about communication or global observability. Another advantage of DecRSPI is that it focuses the computation only on reachable states. As the experiments show, little sampling is needed for domains where agents have sparse interaction structures, and the solution quality calculated by a small set of samples is quite close to the best existing planning algorithms. Most importantly, DecRSPI has linear time complexity over the number of agents. Therefore DecRSPI can solve problems with up to 20 agents as shown in the experiments. Additionally, DecRSPI bounds memory usage as other MBDP-based algorithms. In the future, we plan to further exploit the interaction structure of agents and make even better use of samples, which will be helpful for large real-world domains.

## Acknowledgments

This work was supported in part by the Air Force Office of Scientific Research under Grant No. FA9550-08-1-0181, the National Science Foundation under Grant No. IIS-0812149, the Natural Science Foundations of China under Grant No. 60745002, and the National Hi-Tech Project of China under Grant No. 2008AA01Z150.